\documentclass{article}

\usepackage{PRIMEarxiv}

\usepackage[utf8]{inputenc} 
\usepackage[T1]{fontenc}    
\usepackage{hyperref}       
\usepackage{url}            
\usepackage{booktabs}       
\usepackage{amsfonts}       
\usepackage{nicefrac}       
\usepackage{microtype}      
\usepackage{lipsum}
\usepackage{graphicx}
\graphicspath{{media/}}     
\usepackage{multirow} 
\usepackage{tabularx}

\usepackage[utf8]{inputenc} 
\usepackage[T1]{fontenc}    
\usepackage{hyperref}       
\hypersetup{
    colorlinks=true,
    linkcolor=blue,
    filecolor=magenta,      
    urlcolor=cyan,
    pdftitle={Overleaf Example},
    pdfpagemode=FullScreen,
    }
\usepackage{url}            
\usepackage{booktabs}       
\usepackage{amsfonts}       
\usepackage{nicefrac}       
\usepackage{microtype}      
\usepackage{xcolor}         
\usepackage{graphicx, wrapfig}
\usepackage{amsmath}

\DeclareMathOperator*{\argmin}{arg\,min}
\newtheorem{definition}{Definition}[section]
\newcommand\dimension{\mathsf{d}}
\newcommand{\vect}[1]{\mathsf{#1}}
\usepackage{tabularx,multicol,booktabs,multirow,makecell}[width=\textwidth]

\PassOptionsToPackage{options}{natbib}
\usepackage{multirow}
\usepackage[]{mdframed}
\usepackage[font=small,labelfont=bf]{caption}
\title{A prototype-based model for set classification
}

\author{M. Mohammadi  \\
        University of Groningen\\
        The Netherlands\\
        \texttt{mohammadimathstar@gmail.com}\\
        \And
        S. Ghosh \\
        Technical University of Eindhoven\\
        The Netherlands\\
        \texttt{ghosh.sreejita@gmail.com}\\
}

\begin{document}
\maketitle

\begin{abstract}
Classification of sets of input vectors (e.g., images and texts) is an active area of research within both computer vision and natural language processing. A common way to represent a set of vectors is to model them as linear subspaces. In this contribution, we present a prototype-based approach for learning on the manifold formed from such linear subspaces, the Grassmann manifold. Our proposed method learns a set of subspace prototypes capturing the representative characteristics of classes and a set of relevance factors automating the selection of the dimensionality of the subspaces. This leads to a transparent classifier model which presents the computed impact of each input vector on its decision. Through experiments on benchmark image and text datasets, we have demonstrated the efficiency of our proposed classifier, compared to the CNN and transformer-based models in terms of performance, explainability and computational resource requirements.
\end{abstract}
\footnotetext[1]{Preprint. Under review.}

\section{Introduction}

Due to the success of linear subspaces in modelling variations present 
in data sets, they have extensively been 
used by the machine learning (ML) community. In Computer Vision (CV), they have been used as a learning scheme for image-set classification, to model variation stemming from lightning conditions, pose and facial expression(s) within a set of images as illustrated in the works of 
\cite{yamaguchi1998face, fukui2005face, kim2007boosted,  kim2007discriminative, wang2008manifold, wang2009manifold, hamm2008grassmann, harandi2011graph, huang2015projection, wei2020prototype, sogi2020metric}. 
In Natural language processing (NLP), linear subspaces were first used to model variations of topics within a large corpus of texts\cite{deerwester1990indexing}. Later, with the advancement in word-embedding models, they found new applications in modelling texts, such as phrases, sentences and paragraphs \cite{mu2017representing,shimomoto2021text}.
This idea has recently been employed in document classification where a document is considered as a set of words\cite{gong2019document,shimomoto2021text,shimomotosubspace}. These applications demonstrate the capability of subspace representation for set classification where the goal is to assign a label to a set of vectors.

Interestingly, it has also been shown that such linear subspaces form a manifold, called the Grassmann manifold. This has created a mathematical framework to study subspace-based representations and create models that are well-suited for learning on this manifold. However, aforementioned prior works 
suffer from two major drawbacks. First, their reliance on strategies such as the 'Nearest Neighbor' or 'kernel-trick' limits their applications to small data sets. Second, their performances are highly sensitive, dependent on the hyperparameter values. 
These limitations necessitate the development of more effective approaches for learning on the Grassmann manifold. 

The Generalized Learning Vector Quantization (GLVQ) is a family of distance-based classifiers that uses a set of prototypes 
as typical representatives of the 
classes within the data space \cite{sato1995generalized}. They classify a sample based on its proximity to the learned prototypes, following the Nearest Prototype Classification (NPC) strategy.  
To increase its flexibility and interpretability, several variants of GLVQ have been proposed with different adaptive distances (or dissimilarities) introducing different levels of complexity and transparency such as, \cite{Hammer2005, schneider2009}. This leads to the application of GLVQ in different areas where interpretability is a must \cite{holzinger2017we, arrieta2020explainable, doshi2017towards, bibal2016interpretability}, 
from anthropocentric sectors like healthcare \cite{ghosh2020visualisation, ghosh2025interpretable,hankel2017sequence} and education \cite{widiantara2023application}, to astronomy \cite{mohammadi2022detection}. 

Recently, an extension of GLVQ called the Generalized Relevance Learning Grassmann Quantization (GRLGQ)\cite{mohammadi2024generalized}, addresses the aforementioned challenges of subspace-based learning. 
GRLGQ achieved this through the introduction of a set of subspace prototypes, which summarizes major characteristics of classes, and an adaptive distance, which addresses the adverse effect of dimensionality $\dimension$ on the performance of the algorithm. 
This however comes by partially trading off 
GLVQ's inherent explainability. Specifically, it struggles to precisely quantify the effect of input vectors on the final prediction. 

Therefore, in this paper we introduce our newly developed transparent model that elucidates the contributions of individual input elements (local regions of an image or words) to the model's prediction, while still retaining 
the advantages of GRLGQ. 
The major contributions of this paper are as follows: 
    (1) We extend the family of GLVQ classifiers by introducing a new adaptive distance measure. 
    (2) We demonstrate how the new model explains its decision, thus addressing the aforementioned limitation of GRLGQ. 
     Finally, through experiments on nine benchmarks, (3) we illustrate its performance and interpretability concerning both image and text data.

This paper is organized as follows: 
In section \ref{sec:background}, we provide the necessary background information, including a short introduction about the Grassmann manifold and the problem setting.
In section \ref{sec:method}, we first present our proposed model, and then
we show how it provides details about its decisions.
In Section \ref{sec:feature_extraction}, we describe the feature extraction process, where we use Convolutional Neural Networks (CNN) for images and word embedding models for texts to obtain meaningful representations.
Section \ref{sec:experiment} contains the application of the new method to several benchmarks, including both images and texts.
Finally, we conclude our work in section \ref{sec:conclusion}.

\section{Background} \label{sec:background}

\subsection{Grassmann manifold}

The extensive usage of linear subspaces in different fields has necessitated the study of mathematical structures driven by them, such as the Grassmann manifold. 

\begin{definition}
    The Grassmann manifold $\mathcal{G}(D, \dimension)$ consists of all $\dimension$-dimensional subspaces in $\mathbb{R}^D$. 
\end{definition}

A conventional representation of a point (i.e., subspace) on the Grassmann manifold involves using an orthonormal matrix $P \in \mathbb{R}^{D \times \dimension}$, where its columns serve as a basis for the subspace. Notably, this representation is not unique, meaning any orthonormal matrix spanning the subspace can be employed for representation. It has been shown that two orthonormal matrices $P_1$ and $P_2$ span the same subspace if and only if there exist rotations matrices $Q_1, Q_2 \in \mathcal{O}(\dimension)$\footnote{$\mathcal{O}(\dimension)$: the group of $\dimension \times \dimension$ orthonormal matrices.} such that $P_1 Q_1 = P_2 Q_2$. In other words, different representations of a point on the Grassmann manifold can be transformed into each other using a pair of orthonormal matrices. 

In order to perform a classification task on the Grassmann manifold, it is necessary to have a measure that assesses the (dis)similarity among points on the manifold. Various measures based on canonical correlation (and principal angles) have been proposed. Canonical correlation can be seen as a generalization of the cosine similarity measure to $\dimension$-dimensional subspaces, allowing for the comparison of $\dimension$ pairs of vectors instead of just one.

\begin{definition}{(Canonical correlation)} \label{def:cc} 
Let us consider two subspaces $\mathcal{L}_1$ and $\mathcal{L}_2$ with the dimensionalities of $\dimension_1$ and $\dimension_2$, respectively. Then, $\dimension$ \emph{principal angles}
\[
0 \leq \theta_1 \leq \theta_2 \leq \cdots \leq \theta_\dimension \leq \frac{\pi}{2} \enspace, \hspace{0.4cm} \dimension = \min{ \{\dimension_1, \dimension_2\} }
\]
are defined recursively by the following conditions:
  \begin{align}
      \label{eq:canonicalcorr}
      \cos{\theta_i} &= \max_{\substack{\vect{u}_i^{\prime} \in \mathcal{L}_1,\\ \vect{v}_i^{\prime} \in \mathcal{L}_2}} \vect{u}^{\prime T}_i \vect{v}^{\prime}_i = \vect{u}_i^T \vect{v}_i, \text{ s.t. }
    \begin{cases}
    \lVert \vect{u}_i^{\prime} \rVert = \lVert \vect{v}_i^{\prime} \rVert = 1,&\\
    \vect{u}_i^{\prime T} \vect{u}_j^{\prime}=\vect{v}_i^{\prime  T} \vect{v}_j^{\prime}=0, &\forall j\! < i\!,
    \end{cases}\forall i\!=[1, 2, \cdots, \dimension]
    \end{align}
   Then, the cosine of the principal angles is referred to as the \emph{canonical correlation}, and the elements of  $\{ \vect{u}_k \}_{k=1}^\dimension$ and $\{ \vect{v}_k \}_{k=1}^\dimension$ are known as \emph{principal (or canonical) vectors}\cite{zhang2018grassmannian}.
\end{definition}
Simply put, a pair of principal vectors, denoted by $(\vect{u}_i, \vect{v}_i)$, consist of unit vectors (from $\mathcal{L}_1$ and $\mathcal{L}_2$, respectively) with the smallest angle between them, while also being orthogonal to previously paired vectors. A common method for computing principal angles and vectors is through Singular Value Decomposition (SVD). Given two orthonormal matrices $P_1$ and $P_2$, spanning subspaces $\mathcal{L}_1$ and $\mathcal{L}_2$ respectively, the SVD of $P_1^T P_2$ provides us with pairs of principal vectors and their corresponding canonical correlations\cite{bjorck1973numerical}:
\begin{equation}
    \label{eq:compute_cc}
    P_1^T P_2 = Q_{1,2} (\cos{\Theta}) Q_{2,1}^T \enspace,
\end{equation}
where $\cos{\Theta}$ is a diagonal matrix with canonical correlations, and rotation matrices fulfill the condition: $Q_{1,2}Q^T_{1,2} = Q^T_{1,2}Q_{1,2} = Q_{2,1} Q_{2,1}^T = Q_{2,1}^T Q_{2,1} = I$. Similarly, the principal vectors are given by:
\begin{equation}
    \label{eq:PQ}
	U = P_1 Q_{1,2} = [\vect{u}_1, \cdots, \vect{u}_{\dimension}], \hspace{0.5cm}
	V = P_2 Q_{2,1} = [\vect{v}_1, \cdots, \vect{v}_{\dimension}]\enspace.
\end{equation}

Based on 
canonical correlations, several distances have been proposed including the following:

\begin{definition}
Let $\mathcal{L}_1, ~\mathcal{L}_2$ be on the manifold $\mathcal{G} (D, \dimension)$. Then, the chordal distance is defined as: 
\begin{align}
    \label{eq:geo_dist}
    d_c(\mathcal{L}_1, \mathcal{L}_2) &= \lVert \sin{\Theta} \rVert_2 = \left( \sum_{i=1}^{\dimension} \sin^2{\theta_i} \right)^{1/2} = \left( \dimension - \sum_{i=1}^{\dimension} \cos^2{\theta_i} \right)^{1/2} \enspace.
\end{align}
\end{definition}

\begin{figure}
    \centering       
    \includegraphics[width=\textwidth]{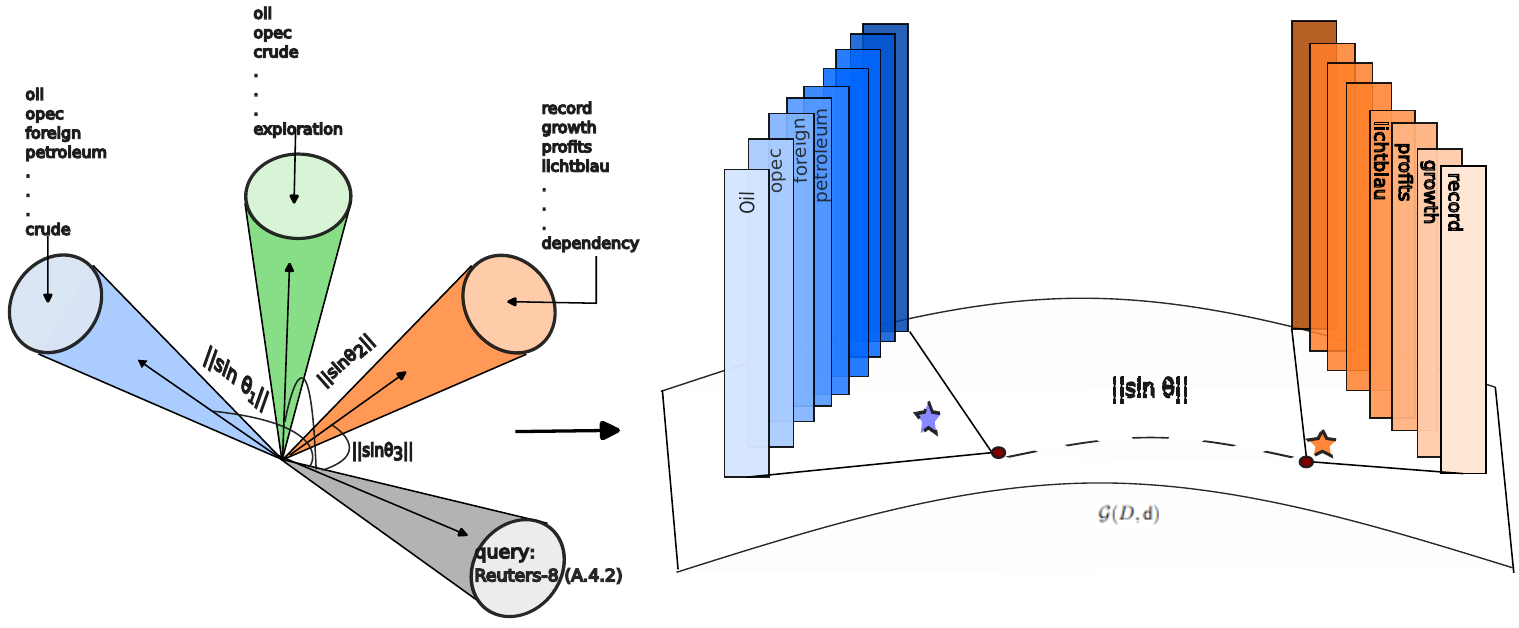}
    \caption{Left) Chordal distance between a query and texts from the Reuters-8 dataset. Right) A schematic figure displaying each text as a point on the Grassmann manifold $\mathcal{G}(D, d)$. Note that colors encode labels of each text.}
    \label{fig:reuters-8-manifold}
\end{figure}
The interpretability of the variant of GLVQ we introduce in this contribution, is due to the exploitation of Chordal distance.

\subsection{Problem setting - set classification} \label{subsec:problem_setting}

We assume a data point as a collection of vectors, denoted by $\{\vect{x}_1, \vect{x}_2, \cdots, \vect{x}_n\}$, as is common in both natural language processing (NLP) and computer vision (CV). The classifier's task is to assign an appropriate label to it.

In NLP, documents or texts are often represented as sets of word embeddings. For example, a text $g = (w_1, w_2, \cdots, w_n)$ of length $n$, where $w_i$ is the $i^{th}$ word, can be transformed using a word embedding model that assigns a vector $\vect{x}_i \in \mathbb{R}^D$ to each word $w_i$.
By stacking these word vectors, the text can be represented in matrix form: $X = [\vect{x}_1, \vect{x}_2, \cdots, \vect{x}_n]$. 
However, this representation is still incapable of addressing the challenge of comparing sets of variable lengths.

In CV, it is common to extract multiple feature vectors from a single image to capture various aspects of its content. For example, techniques such as the Bag-of-Visual-Words model\cite{csurka2004visual} involve detecting and extracting local features (e.g. SIFT\cite{lowe2004distinctive}  descriptors) from an image, resulting in a set of feature vectors that describe different regions or characteristics of the image. 
Image set classification goes beyond analyzing individual images; it involves analyzing collections of images that depict various instances or conditions of an object. This is valuable in applications such as as face recognition, where multiple images of the same individual in different poses, lighting, or expressions provide a complete representation, enhancing the recognition accuracy and robustness.

Representing these sets of feature vectors in a consistent and comparable manner is crucial for effective classification. A possible solution is to assume $X$ as a low-rank matrix represented by its SVD as:
\begin{equation}
\label{eq:eigendecomposition}
X = P S R^{T} \enspace,
\end{equation}
where $S_i$ is a ($\dimension \times \dimension$) diagonal matrix containing singular values in descending order, and columns of $P$ and $R$ (of size $D \times \dimension$) are left and right singular vectors, respectively.
The columns of $P$ form the basis for the subspace generated by the columns of $X$. This leads to a modified training set: 
\[
\{ (P_i, y_i) ~\vert ~i= 1, \cdots, N\}, 
\]
where $P_i \in \mathbb{R}^{D \times \dimension}$ is an orthonormal matrix. 
As this representation implies the assumption that all training examples reside on the Grassmann manifold $\mathcal{G}(D, \dimension)$, only a  
classifier that operates on this manifold can effectively learn from the data.

\section{Method} \label{sec:method}

In this section, we present an overview of our novel contribution, which builds upon the foundations of GRLGQ via the incorporation of a distinct dissimilarity measure to enhance its explainability. Subsequently, 
we demonstrate how the resulting model offers insights into their decision-making processes, providing details about the role of each input in its predictions.

\subsection{Adaptive Chordal Distance and Subspace-based LVQ (AChorDS-LVQ)} \label{subsec:grlgq}

Let $\{(P_i, y_i)\}_{i=1}^N$ be the training set on the Grassmann manifold $\mathcal{G}(D, \dimension)$. Following the GLVQ algorithm, we use a set of labelled prototypes $\{ (W_i, c_i) \}_{i=1}^p$ to model the distribution of classes on the data space (i.e. Grassmann manifold). 
Here, prototypes are orthonormal matrices on $\mathcal{G}(D, \dimension)$, representing the corresponding classes by capturing their respective major characteristics. 
Learning takes place by minimization of the cost-function Eq.\ (\ref{eq:cost_function}). 
\begin{align}
    \label{eq:cost_function}
    E ( \mathcal{W} )=& \sum_{i=1}^n \phi \left( \frac{d_c(P_i, W^+) - d_c(P_i, W^-)}{d_c(P_i, W^+) + d_c(P_i, W^-)} \right)  \text{ where, } \\
    \label{eq:adaptive_chordal}
    d_c(P, W) =& 1 - \sum_{k=1}^{\dimension} \lambda_k \cos{\theta_k} = 1 - \sum_{k=1}^{\dimension} \lambda_k \vect{u}_k^T \vect{v}_k, \hspace{0.3cm} \sum_{i=1}^{\dimension} \lambda_i = 1 \enspace,
\end{align}
where $d_c(P, W) $ is the adaptive distance between a prototype and a data point, 
$\{ \lambda_i \}_{i=1}^{\dimension}$ are the relevance factors capturing the contribution of the canonical correlations towards differentiating the classes,  
and $\mathcal{W} = \{ W_1, \cdots, W_p\}$ is the set of all prototypes. 
Note that $W^+$ and $W^-$ are the closest prototypes to $P_i$ with the same and different labels, respectively. 
Moreover, $\phi$ is a monotonically increasing function.\footnote{Sigmoid and identity functions are two common choices for $\phi$.} As the definition of $d_c$ is motivated from the chordal distance, we refer to our newly introduced method hereafter as 
\textbf{A}daptive \textbf{Chor}dal \textbf{D}istance and \textbf{S}ubspace-based LVQ (AChorDS-LVQ for short).

During the training process, we use the Stochastic Gradient Descent (SGD) algorithm to optimize the cost-function (Eq.\ \eqref{eq:cost_function}) and find the ideal locations of the prototypes on the manifold and the optimum values of the relevance factors.  
Details about the computation of gradients can be found in Appendix \ref{subsec:gradient-computation}.
In the prediction phase, it follows the NPC strategy to 
assign the sample 
the label of the 
prototype nearest to it:
\begin{align}
     \tilde{c}(X) &= c(W_k) \enspace, \text{ s.t. } 
       k = \argmin _{i=1, \cdots, p} d( X, W_i ) \enspace, 
     \label{eq:NearesetProto}
\end{align}
where $\tilde{c}(X)$ is the predicted label of $X$. Note that as the number of prototypes $p$ is much smaller than the number of 
instances $N$, NPC strategy provides a computational- and memory- cost-efficient model during the test phase,                           compared to other classifiers following the Nearest Neighbor strategy or those using kernel trick.

\subsection{Explainability} \label{subsec:explainability}
With the increasing real-world applications 
of ML models, performance metrics alone do not suffice for reliability, 
highlighting the critical need for model interpretability and explainability \cite{arrieta2020explainable, doshi2017towards, ribeiro2016model}. Even though model explainability is demanded in anthropocentric applications, it is important in other sectors as well to identify the sources of uncertainties and noise in the model training and deployment pipelines, and strategize appropriately to address these \cite{ribeiro2016model}, improve robustness, minimize algorithmic bias and ensure fair and responsible use of AI\cite{holzinger2017we, arrieta2020explainable, slack2020fooling}. Local interpretable model-agnostic explanations (LIME) \cite{ribeiro2016should} and (SHapley Additive exPlanations) (SHAP)\cite{lundberg2017unified} are among the most widely used XAI tools which provide post-hoc explanations of decisions by even complex black-box models, such as any model with a deep neural network (DNN) architecture \cite{slack2020fooling}. \cite{slack2020fooling} also demonstrate the susceptibility of these XAI techniques, which function by (i) perturbing the region close to a sample-of-interest and (ii) applying simpler (often linear) surrogate models, to being fooled by adversarial classifiers, thereby misleading model designers, end-users and stakeholders resulting in severe ramifications \cite{borrego2022explainable}. Intrinsically interpretable ML models (e.g., linear and logistic regressors, decision trees (DT)s, and NPCs) 
circumvent these risks since they provide direct access to their respective decision-making logic. In this section, we elucidate what makes AChorDS-LVQ intrinsically interpretable.

AChorDS-LVQ performs adaptive distance-based classification following the NPC strategy. 
Thus, one can measure the influence of different input vectors on the  
decisions of AChorDS-LVQ by evaluating their impact on the 
measured effective proximity between the sample and the prototypes. 
Here we show how the influence of the input vectors is exactly computed with the newly introduced Chordal distance. 
Let $X$ be a matrix containing a set of vectors, i.e.:
\[
X = [\vect{x}_1, \vect{x}_2, \cdots , \vect{x}_n]
\]
Using the SVD, one can derive its corresponding $\dimension$-dimensional subspace:
\[
P: X = P S R^T \enspace,
\]
where columns of $P$ contain the basis for the generated subspace.
To predict $P$'s label, the model calculates its distances to all prototypes $W$, by computing the 
principal vectors $U$ corresponding to each pair (following Eq.\ \eqref{eq:compute_cc}, \eqref{eq:PQ}):
\[
U = PQ_P = X M
\]
where $M_{n \times \dimension}=R S^{-1} Q_{P}$. 
Let $\vect{m}_i$ denote $i^{th}$ columns of $M$. Then, we have $\vect{u}_i = X \vect{m}_i$, where elements $\vect{m}_i$ capture the contribution of input vectors (i.e., columns of X) on the construction of the $i^{th}$ principal vectors. Additionally, we define $\mathrm{x}_j$ as the $j^{th}$ row of $X$ which is the $j^{th}$ coordinate of the input vectors, resulting in: 
\begin{align}
u_i^{j} = \mathrm{x}_j \vect{m}_i = \sum_{k=1}^n x_j^k m_i^k, 
\end{align}
where $u_i^{j}$ is $j^{th}$ coordinate of $i^{th}$ principal vector $\vect{u}_i$. In Eq.\ \eqref{eq:pixel_rank}, the right-hand term within
\begin{align}
    d_c(P, W) &= 1 - \sum_{i=1}^{\dimension} \lambda_i \vect{u}_i^T \vect{v}_i =  1 - \sum_{i=1}^{\dimension} \sum_{j=1}^D \lambda_i u_i^j v_i^j \nonumber \\
    &= 1 - \sum_{k=1}^n \Biggl[ \sum_{j=1}^D  \Biggl( \sum_{i=1}^{\dimension} \lambda_i  m_i^k v_i^j \Biggr) x_j^k \Biggr]
    \label{eq:pixel_rank}    
\end{align}
parenthesis represents the impact of the $j^{th}$ element of the $k^{th}$ input vector on the computed distance. Consequently, this allows for these elements to be ranked 
according to their influence on this distance, and enabling more precise estimation of possible sources of uncertainty and noise, among others. The terms within the square brackets in Eq.\ \eqref{eq:pixel_rank}, i.e., the cumulative sum of elements of an input vector $\vect{x}_k$ (for $k=1, \cdots, n$), reflect the overall impact of this 
input vector on the computed distance between the input set $X$ and a prototype. 

To decide about an input set $X$, the model computes its distance to the nearest (winner) prototypes $W^{\pm}$. Thus, as formulated in the numerator of the cost-function Eq.\ \eqref{eq:cost_function}, we use the difference between these distances to find the most influential elements for the input set's prediction:
\begin{equation}
    d_c(P, W^-) - d_c(P, W^+) = 
\sum_{k=1}^n \Biggl[ \sum_{j=1}^D \biggl( \sum_{i=1}^d \lambda_i  m_i^k (v_i^{j+} - v_i^{j-}) \biggr) x_j^k \Biggr] \enspace,    
    \label{eq:explainability_diff}
\end{equation}
Thus, Eq.\ \eqref{eq:explainability_diff} illustrates how AChorDS-LVQ can define with element-level precision, the impact of the input vectors 
on the final prediction. 

\section{Feature Extraction and Representation}\label{sec:feature_extraction}

In this section, we discuss how features are extracted from images (leveraging the convolutional neural networks (CNN)) and how text data is transformed into numerical representations (word embedding) to be used for classification. 
These techniques enable us to process complex unstructured data and make them 
suitable for machine learning models.

\subsection{Image Feature Extraction}
\begin{figure}
    \centering
    \includegraphics[width=1\linewidth]{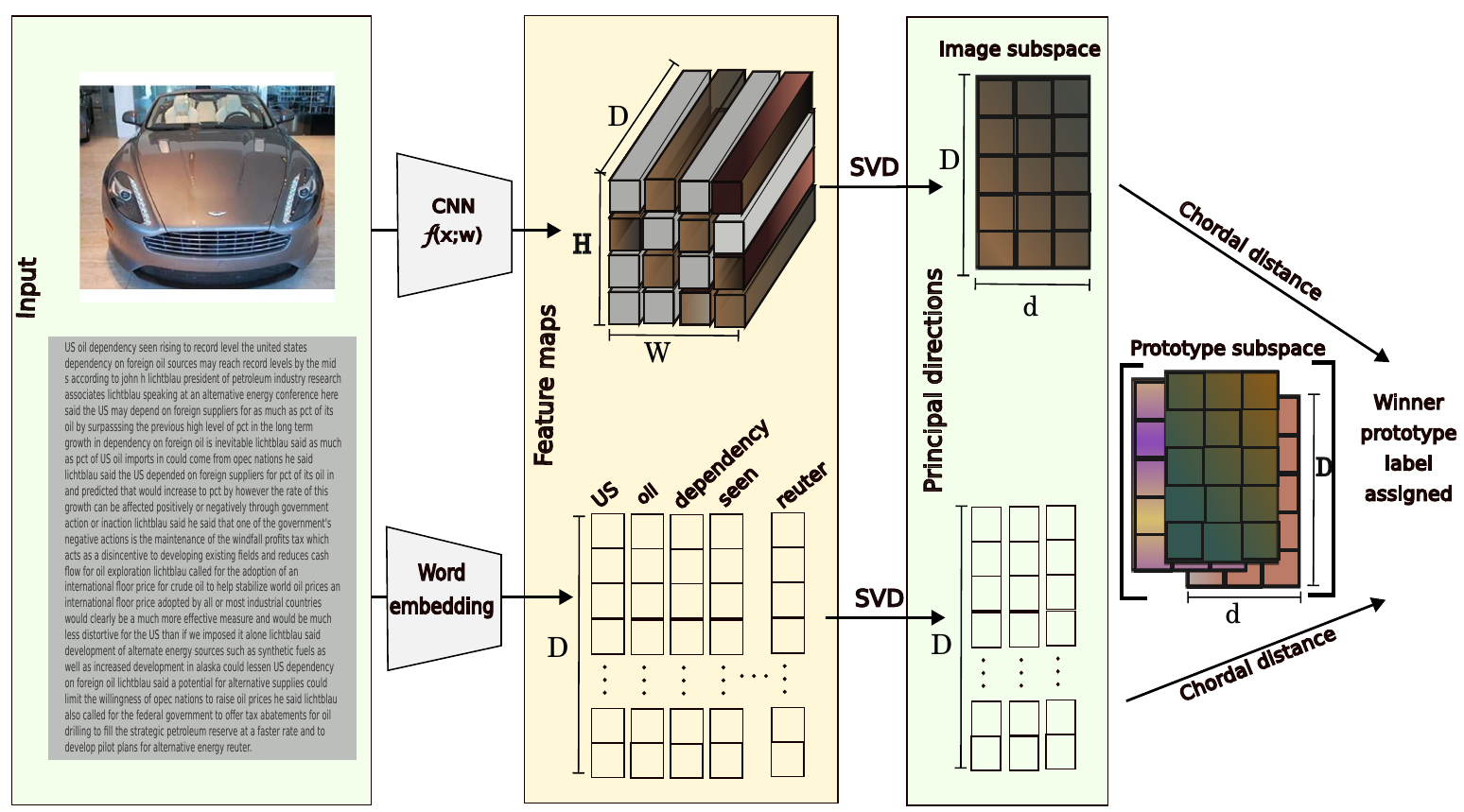}
    \caption{The architecture of the application of AChorDS-LVQ: Above) CNN as the backbone network for image classification, Bottom) Word Embedder model for text classification. In both cases, the SVD is applied after the feature extraction and. Thereafter, the distance introduced in \ref{eq:adaptive_chordal} is used to find the winner prototype and assign its label to the input sample.}
    \label{fig:feature_engineering}
\end{figure}

Effective image classification requires the extraction of robust and discriminative features. With the rise of CNNs in the early 2010s, there was a shift from handcrafted features like SIFT, HOG, and SURF to data-driven learned features, leading to significant improvements in complex computer vision problems.

Building upon this advancement, we integrate AChorDS-LVQ with a CNN backbone architecture. As depicted in Figure \ref{fig:feature_engineering} above, an input image is first processed through the CNN function $f$. This produces a convolutional output $z = f(x; w)$ where $w$ is the trainable parameters of $f$. This output consists of ($W \times H$) $D$-dimensional feature vectors, corresponding to different regions of the image.

Subsequently, we apply the SVD to this collection of $(W \times H)$ feature vectors, treating them as a set without considering their specific spatial positions within the image. This approach allows the model to focus on the presence of patterns regardless of their locations, enabling the detection of features that can appear anywhere in the image. The SVD gives a $\dimension$-dimensional subspace in $\mathbb{R}^D$, which serves as the image's representation. 
This subspace is then compared with prototypes using the distance metric introduced in Equation \ref{eq:adaptive_chordal}, and the image is assigned the label of the closest prototype.

Integrating AChorDS-LVQ with a CNN backbone architecture thus allows for the construction of prototypes directly from the latent representations of images. This approach combines the feature extraction capabilities of CNNs with the prototype-based classification strengths of AChorDS-LVQ, aiming to enhance both performance and interpretability in image classification tasks. 
Note that since each feature vector represents a specific region of the image, we can derive a score indicating the importance of these regions, using Eq. \eqref{eq:explainability_diff}. To obtain pixel-wise importance 
we upsample these region-based scores using bicubic interpolation, as done in \cite{nauta2021neural}, aligning the scores with the original input shape. This enables us to highlight and visualize the most influential pixels in the image, offering valuable insights into the model’s decision-making process.

\subsection{Word embedding}

In text classification, we can consider a document as a set of words, each contributing uniquely to its meaning. To process textual data effectively, it is essential to convert these words into numerical representations that capture their semantic relationships. In simpler terms, if two words have similar meanings, their numerical representations should also be close to each other. This need led to the development of several word embedding models, such as Word2Vec \cite{mikolov2013efficient} and GloVe \cite{pennington2014glove}, which determine word similarity based on their co-occurrence patterns. 
While Word2Vec and GloVe effectively encode semantic relationships, they treat words in isolation, lacking sensitivity to varying contexts. Transformer-based models, such as BERT (Bidirectional Encoder Representations from Transformers) \cite{devlin2018bert}, address this limitation by generating context-dependent embeddings. In other words, whereas models like Word2Vec and GloVe assign a fixed vector to each word, transformer-based approaches generate dynamic embeddings, allowing the same word to have different representations depending on its context.

As shown in Figure \ref{fig:feature_engineering} bottom, here a word embedding model is used to generate word embedding vectors for the words in the text. Thereafter, 
we apply 
SVD to the resulting matrix $X = [v_1, v_2, ..., v_n]$ of embedding model, where $v_i$ represents the embedding vector for the $i^{th}$ word. This process allows us to obtain a subspace representation for each text, which can then be used to train AChorDS-LVQ. Note that the resulting prototypes will also be represented as subspaces in the embedding space.

\section{Experiment} \label{sec:experiment}

In this section, we present an evaluation of our proposed algorithm by comparing its performance with existing methods across both text and image classification tasks. First, we focus on text classification and evaluate our approach on three benchmark datasets, comparing it with various methods, including transformer-based models such as BERT and RoBERTa. Next, we assess its performance in image classification by benchmarking it against previous prototype-based approaches on three widely used datasets. Finally, we extend our evaluation to image-set classification, where a set of images represents different variations of an object. 
Through these experiments, we aim to showcase the robustness, accuracy, and explainability of our method across diverse classification tasks.

\subsection{Document classification}
To evaluate the performance of the proposed approach for document classification, we applied it to three document classification tasks across a range of document lengths: (a) two datasets with shorter texts, i.e., Reuters-8\cite{cachopo2007improving} and Hyperpartisan\cite{kiesel-etal-2019-semeval}, (b) Arxiv-4\cite{liu2018long} which is a dataset with long texts (with an average length of 6000 words). In our experiments, we use the GloVe and the Word2Vec models for generating word embeddings, due to their flexibility across varying document lengths. To prevent noisy influence of stop words we remove these during preprocessing.  
Additionally, we set $\dimension=20$ for the first two experiments and $\dimension=30$ for the last one (with longer texts). 

To assess the performance of our proposed approach, we compared the performance of our proposed model against those including the traditional yet widely-used Support Vector Machine (SVM), subspace-based models: MSM\cite{shimomoto2021text} and GRLGQ, and two transformer-based models: BERT and RoBERTa. For the Arxiv-4 dataset containing longer texts, we additionally consider the performance of algorithms proposed in \cite{beltagy2020longformer,liu2018long, he2019long} for handling long texts. 
For models that are not transformer-based, word embedding models, such as Word2Vec and GloVe have been used for generating vector representations of texts, as specified in \autoref{tab:reuters-8}. To further explore the influence of different embedding methods on our model's efficacy, we utilized Word2Vec, GloVe, and BERT embeddings. Finally, given the computational cost of transformer-based models, we compared two common practices for their fine-tuning: updating all parameters as well as LoRA \cite{hu2021lora}, a method to reduce training time. 

\begin{table}[]
    \centering
    \begin{tabular}{cccccc}
        \hline
        \textbf{Methods} & \makecell{\textbf{Word}\\ \textbf{Emb.}} &
        \makecell{$N_{params}$\\ (train)} &\textbf{Reuters8} & \makecell{\textbf{Hyper-}\\ \textbf{partisan}} & \textbf{Arxiv4}\\
        \toprule
        MSM\cite{shimomoto2021text} &W2V& & &92.01 & -\\  \hline         
         SVM &GloVe &457k & 96.03 & 70.77 & 93.64\\
         \hline
         \multirow{4}{*}{GRLGQ} &W2V&48k& 97.67 &90.77& 96.06\\
         &GloVe& 48k& \makecell{97.93\\($\pm 0.09$)} & \makecell{91.28\\ ($\pm 0.73$)} & \makecell{96.65\\ ($\pm 0.04$)} \\
         &BERT& 184k& \textbf{98.17} & 90.77& -\\
         \midrule
         \multirow{4}{*}{\makecell{AChorDS-\\ LVQ}} & W2V &48k& 97.94& 90.77 &  96.06\\ 
         &GloVe &48k& \makecell{97.95\\($\pm 0.07$)}& \makecell{\textbf{91.80}\\ ($\pm 0.73$)} & \makecell{\textbf{96.74}\\ ($\pm0.17$)} \\ 
         &BERT &184k& \textbf{98.17}&90.77 & - \\
         \midrule
         $\text{BERT}^\text{FULL}$& & 110 M & 98.04& 90.77& 96.35\\
         $\text{BERT}^\text{LoRA}$& &300k   & 97.94& 90.77& - \\
         \midrule
        $\text{RoBERTa}^\text{FULL}$ & &125M& 98.08& 90.77& 95.45 \\ 
        $\text{RoBERTa}^\text{LoRA}$& &892k& 98.08& 90.77& - \\
         \midrule
         Longformer& &-& -& -& 95.12 \\
         \midrule
        \multirow{1}{*}{CNN+*+Agg\cite{liu2018long}}& & & & & \\         
        $\bullet$ Rand &  &- & -& -& 94.02\\
         $\bullet$ LSTM &  &- &- &- & 94.25\\
         $\bullet$ RAM & & & & & 94.73\\  \midrule
         Glimpse\cite{he2019long} & & & & & 94.18 \\
         \bottomrule
    \end{tabular}
    \caption{The performance of different classifiers on the document classification task.}
    \label{tab:reuters-8}
\end{table}

\begin{figure}
    \centering
    \includegraphics[scale=1.5]{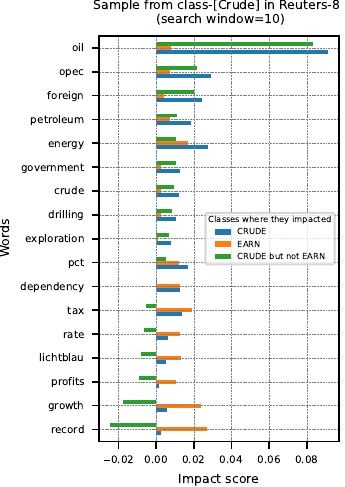}
    \captionof{figure}{Words with the highest impact on the prediction of a sample from \emph{Crude} class of the Reuters-8 dataset.}
    \label{fig:topwords_retuers-8}
\end{figure}

\paragraph{Performance:} The results for the three text-based datasets are 
presented under their respective columns in \autoref{tab:reuters-8} which demonstrates the competitive performance of AChorDS-LVQ. 
Specifically, the combination of GloVe embeddings and AChorDS-LVQ demonstrates comparable performance to transformer-based models for the Reuters-8 dataset and outperforms them for the Hyperpartisan and Arxiv-4 datasets. This is achieved with significantly fewer parameters (48k) compared to BERT (with 110M for full fine-tuning and 300k with LoRA) and RoBERTa (with 125M for full fine-tuning and 892k with LoRA). 
Given the large number of parameters in transformer-based models, we needed GPU for accelerated computations during training. However, for training AChorDS-LVQ, CPU was enough (\autoref{A-sec:Compute resource}), which makes it computationally efficient. 

\paragraph{Explainability:} Following our discussion in \autoref{subsec:explainability}, we can use 
Eq.\ \eqref{eq:explainability_diff} to determine the impact of each word on the model's decisions, and rank them accordingly. 
\autoref{fig:topwords_retuers-8} provides the influential words for a text from `crude' class. The blue and orange bars corresponding to each term indicate the extent of impact of these words from the selected document, in driving the model's decision towards class `crude', representing the crude oil market, and class `earn', representing the topic of earning through investments respectively. The green bar represents the net impact of this term in distinguishing 
`crude' from `earn' (Eq.\ \eqref{eq:explainability_diff}); the negative direction of the green bar indicate that the corresponding terms are more representative of the closest incorrect class (here `earn'). This also shows that terms related to the `crude oil' market, such as `oil', `opec', `crude', `energy', `petroleum' and `Premier Composite Technologies (PCT)', a leading global supplier, are correctly identified from the document by our model, as drivers of its decisions about this document. Further, the terms whose corresponding green bar is obscure indicate potential sources of uncertainties for the model (e.g., `dependency'), as these are equally relevant for both these classes and hence not helpful in differentiating them. 
Remarkably, despite some words being associated with the oil market, such as `crude', `opec' and `petroleum', appearing only once or twice, the model accurately identifies them as important, reflecting their significance in capturing key class characteristics for identifying the topic `crude oil'.\footnote{\autoref{fig:reuters_example} in appendix contains the full sample text corresponding to \autoref{fig:reuters_example}.} This illustrates that the learned prototypes of our model have successfully encapsulated the general topic characteristics and keywords of each class. \autoref{A-sec:Explain} contains further examples of such explanations for other samples from `Earn' class, and samples from the Hyperpartisan and Arxiv-4 datasets.

\subsection{Image classification}

To evaluate the effectiveness of AChorDS-LVQ in image classification, we integrated it with two backbone networks: ResNet-50 and ConvNeXt-Tiny. We then tested these configurations across three benchmark datasets: a) the CUB-200-2011 dataset, which consists of 200 bird species and provides a comprehensive set of fine-grained categories; b) the Stanford Cars dataset, containing 196 classes of different car models, presenting challenges in distinguishing between similar vehicle types; and c) the Oxford Pets dataset, which includes 37 categories of cats and dogs. By employing both ResNet-50 and ConvNeXt-Tiny as backbone networks within the AChorDS-LVQ framework, we aimed to assess the model's adaptability and performance across these diverse datasets. 
The methods we compare to are prototype-based models such as ProtoPool\cite{rymarczyk2022interpretable}, Prototype Trees (ProtoTree)\cite{nauta2021neural}, Patch-Based Intuitive Prototypes Network (PIPNet)\cite{nauta2023pip}, prototype-based network (PBN) for Classification-by-Components (CBC)\cite{saralajew2024robust}, and prototype-based Vision Transformers (ProtoViT)\cite{ma2024interpretable}.

\begin{table}
    \centering
    \begin{tabular}{cc|c c c}
        &\textbf{Methods} & \textbf{CUB200} & \textbf{CAR} & \textbf{PET} \\
        \hline
        \multirow{4}{*}{\rotatebox[origin=c]{90}{ResNet}}&ProtoPool\cite{rymarczyk2022interpretable} & 85.5 ($\pm 0.1$)& 88.9 ($\pm 0.1$)& - \\
        &ProtoTree\cite{nauta2021neural} & 82.2 ($\pm 0.7$)& 86.6 ($\pm 0.2$)& -\\
        &PIPNet\cite{nauta2023pip} &82.0 ($\pm 0.3$)& 86.5 ($\pm 0.3$)& 88.5 ($\pm 0.2$) \\
        &PBN-CBC\cite{saralajew2024robust} & 83.3 ($\pm 0.3$) & 92.7 ($\pm 0.1$) & 90.1 ($\pm 0.1$)  \\
        &AChorDS-LVQ & 87.4 ($\pm 0.1$)& 87.9 ($\pm 0.3$)& 91.4 ($\pm 0.0$) \\
        \hline
        \multirow{5}{*}{\rotatebox[origin=c]{90}{ConvNeXt}}&PIPNet\cite{nauta2023pip} & 84.3 ($\pm 0.2$)& 88.2 ($\pm 0.5$)& 92.0 ($\pm 0.3$)\\
        &ProtoPool\cite{rymarczyk2022interpretable} & 85.5 ($\pm 0.1$)& 88.9 ($\pm 0.1$)& 87.2 ($\pm 0.1$)\\
        &ProtoViT\cite{ma2024interpretable} & 85.8 ($\pm 0.2$)& 92.4 ($\pm 0.1$)& 93.3 ($\pm 0.2$) \\
        &PBN-CBC\cite{saralajew2024robust} & \textbf{87.8} ($\pm 0.1$) & 93.0 ($\pm 0.0$) & \textbf{93.9} ($\pm 0.1$)  \\
        &AChorDS-LVQ & 84.9 ($\pm 0.0$) & \textbf{93.2} ($\pm 0.1$)&\textbf{93.9} ($\pm 0.1$) \\
        \hline
    \end{tabular}
    \caption{The accuracies of models for image classification.}
    \label{tab:image_classification}
\end{table}

\paragraph{Performance:} Table \ref{tab:image_classification} presents the performance comparison between AChorDS-LVQ and other prototype-based approaches across the aforementioned datasets. The results highlight the effectiveness of AChorDS-LVQ in various classification tasks, particularly when combined with modern backbone networks.
For the Cars dataset, AChorDS-LVQ with the ConvNeXt-tiny backbone achieved the best performance, demonstrating its ability to extract and utilise meaningful features for fine-grained classification. 
In the Pets dataset, both AChorDS-LVQ and CBC achieved the highest accuracy when paired with the ConvNeXt-tiny backbone, indicating that this combination is particularly well-suited for this dataset.
For the CUB200 dataset, AChorDS-LVQ with the ResNet50 backbone achieved the second-best performance, with an accuracy of 87.4\%, slightly behind PBN-CBC with ConvNeXt-tiny (87.8\%). The difference in performance is relatively small, suggesting that both methods are highly competitive. However, since ResNet50 is generally less computationally demanding than ConvNeXt-tiny, AChorDS-LVQ with ResNet50 could offer a more efficient alternative to PBN-CBC with ConvNeXt-tiny, making it a favorable alternative in scenarios where computational cost is a concern.

The consistent performance of AChorDS-LVQ across multiple datasets underscores its adaptability and effectiveness in diverse image classification scenarios. These findings suggest that the combination of AChorDS-LVQ with advanced backbone networks like ConvNeXt-tiny can lead to significant improvements in classification accuracy.

\begin{figure}
    \centering
    \includegraphics[width=0.3\linewidth]{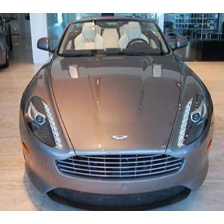}
    \includegraphics[width=0.3\linewidth]{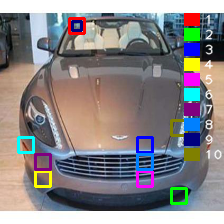}
    \includegraphics[width=0.3\linewidth]{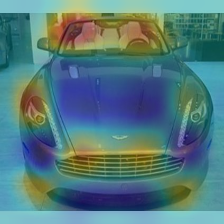}
    \caption{Left) An example image from the Car dataset. Middle) The highlighted regions per principal direction. 
    Right) The combined images comprised both the image and the heatmap generated by our
model.}
    \label{fig:car_example}
\end{figure}

\paragraph{Explainability} AChorDS-LVQ, as a prototype-based approach, provides a natural way to interpret classification results by allowing visualization of the most relevant features used for decision making. To demonstrate the explainability of AChorDS-LVQ, we analyze the classification of an Aston Martin Virage Convertible 2012 from the Cars dataset. 
In figure \ref{fig:car_example} middle, we visualise the closest regions per principal vectors. As can be seen, principal vectors mostly encode different parts of the car. Moreover, figure \ref{fig:car_example} right demonstrates the importance of different pixels in the image as computed by Eq. \eqref{eq:pixel_rank}, using a heatmap overlay. The highlighted areas indicate the most influential features in the model’s decision. The heatmap reveals that the model strongly focuses on the headlights and grille, which align well with how humans distinguish car models. Since the grille is a signature design element for Aston Martin, its prominence in the heatmap suggests that AChorDS-LVQ captures meaningful and discriminative features. Additionally, some activation is observed on the seats through the windshield, 
indicating that the model recognizes higher-level distinguishing attributes beyond just brand and model details. The flood lights, rear-view mirror and emblem also receive moderate attention, suggesting that the model considers shape and design cues in its classification. 
These findings suggest that AChorDS-LVQ effectively leverages human-intuitive features, enhancing its explainability compared to traditional black-box deep learning models. By highlighting the most relevant regions, our approach provides meaningful insights into the classification process, demonstrating that the model focuses on semantically important parts of the image.

\subsection{Image set classification} \label{subsec:image-set-experiment}

In this final experiment, we extend our evaluation from single-image classification to image-set classification, where each sample consists of a set of images rather than a single image. Additionally, unlike previous experiments that relied on CNNs, as feature extractors, here we work directly with raw images without any feature extraction step. We compare our method with earlier works across three datasets: (a) ETH-80 for object recognition, (b) Extended YaleB for face recognition, and (c) UCF for activity recognition. We follow the preprocessing steps described in \cite{wei2020prototype} to ensure fair comparison and reproducibility.

We compare AChorDS-LVQ to other set classifiers that learn on the Grassmann manifold, such as Grassmann Discriminant Analysis (GDA)\cite{hamm2008grassmann}, Graph-embedding GDA\cite{harandi2011graph}, prototype learning and collaborative representation using Grassmann manifolds (GGPLCR)\cite{wei2020prototype}, Projection metric learning on
Grassmann manifold (PML)\cite{huang2015projection}, Discriminative canonical correlations (DCC)\cite{kim2007discriminative}, Affine Hull based Image Set Distance (AHISD)\cite{cevikalp2010face}, Convex Hull based Image Set Distance (CHISD)\cite{cevikalp2010face}, Mutual Subspace Method (MSM)\cite{yamaguchi1998face}, and A-based Metric Learning for
Subspace representation (AMLSL)\cite{sogi2020metric}.

\begin{figure}
    \centering
    \includegraphics[width=0.6\linewidth]{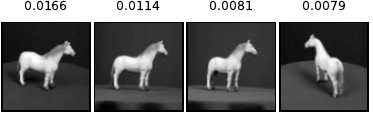}
    \includegraphics[width=0.6\linewidth]{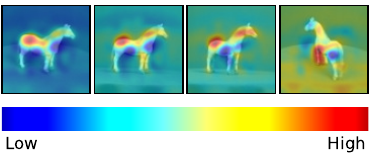}
    \caption{Explainability for ETH-80 dataset:(a) four most important images 
    (impact score above respective image) on prediction; and (b) heatmaps depicting the impact of 
    their constituent pixels.}
    \label{fig:explainability-eth80}
\end{figure}

\begin{table}[]
    \centering
    \begin{tabular}{cccc}
        \toprule
        \textbf{Methods} & \textbf{ETH-80} & \textbf{Ex. Yale B} & \textbf{UCF} \\
        \toprule
        GDA$^*$& $91.00 (\pm 2.13)$ & $82.80 (\pm 0.96)$& 70.00\\
        GGDA$^*$& $92.50 (\pm 1.16)$ & $79.11 (\pm 0.92)$& -\\
        PML$^*$& $93.75 (\pm 1.80)$& $77.08 (\pm 0.18)$& 72.67\\
        AMLSL$^*$ & - & - & 74.00\\
        GGPLCR$^*$& $\mathbf{96.75} (\pm 1.30)$& $88.57 (\pm 0.48)$& -\\
        \midrule
        DCC & $90.75 (\pm 3.34)$ & $77.08 (\pm 3.25)$ & 59.33\\
        MSM & $77.00 (\pm 6.21)$ & $70.00 (\pm 4.12)$ & 48.67\\
        AHISD & $73.00 (\pm 7.24)$ & $55.95 (\pm 5.54)$ & 53.33\\
        CHISD & $75.50 (\pm 4.38)$ & $62.38 (\pm 4.97)$ & 52.00\\
        GRLGQ & $94.50 (\pm 1.87)$& $90.12 (\pm 2.25)$ &  $\mathbf{80.00}$\\
        \midrule
        AChorDS-LVQ  & $94.00 (\pm 2.00)$& $\mathbf{90.18} (\pm 1.64)$ & 79.33\\
        \bottomrule
    \end{tabular}
    \caption{Results (mean accuracy with standard deviations)) of the image-set classification. Methods with (*) are reported from \protect\cite{wei2020prototype} (first two columns.) and \protect\cite{sogi2020metric} (last column).}
    \label{tab:experiment_img}
\end{table}

\paragraph{Performance:} Table \ref{tab:experiment_img} indicates that among the different methods we applied on the aforementioned datasets, GGPLCR, AChorDS-LVQ and GRLGQ achieve the best performances. Among them, GGPLCR has two drawbacks. First, its dependency on the `Nearest Neighbor' prediction strategy directly links its complexity (number of parameters) to the size of the training set. This limits its applicability to small-size data sets. Second, it has been shown in \cite{wei2020prototype} that GGPLCR needs to be tuned with an optimal value for the dimensionality of the subspace $\dimension$. In contrast to GGPLCR, GRLGQ and AChorDS-LVQ overcome these limitations due to their NPC-strategy for prediction. 
Additionally, the relevance factors of AChorDS-LVQ minimise the impact of redundant and noisy dimensions, resulting in a more robust classifier. 

\paragraph{Explainability:} Although GRLGQ and AChorDS-LVQ produce similar results with equivalent complexity levels, their levels of explainability differ. While GRLGQ offers a degree of transparency by highlighting significant pixels and images for each principal angle individually, it fails to specify precise insights into the influence of pixels and images on its final decisions. Conversely, AChorDS-LVQ allows evaluation of the impact of any image on the model's decisions, by computing its influence on distances. Following Eq. \eqref{eq:explainability_diff} (inside square brackets), we can sort images (from an image set) based on their impact on the prediction of the model. The first row in \autoref{fig:explainability-eth80} shows four images (of a horse) with the highest impact. Through the computation of inner parentheses in Eq.\ \eqref{eq:explainability_diff}, we can find the most influential pixels within each image. As seen in the second row of \autoref{fig:explainability-eth80}, the most influential pixels correspond to those representing the mane of the horse, which is a characteristic that differentiates a horse from a dog (the class represented by the closest incorrect prototype). Similarly, the pixels representing the background of the horses have lower relevance than pixels representing the main body, and specifically the characteristic features of the horse. This insight is useful not only to verify the model's decision but also to strategize camera placement for optimal object recognition and exposing possible sources of bias and uncertainty in the input data impacting the model's decision.

\section{Conclusion}\label{sec:conclusion}
In this contribution, we extend the GLVQ family to AChorDS-LVQ by introducing a new distance measure, based on the chordal distance. AChorDS-LVQ assumes that data lies on the Grassmann manifold, and it learns a set of subspace prototypes, encoding the characteristics of classes, and relevance factors, capturing the relevance of canonical correlation for distinguishing the 
classes. In contrast to GRLGQ, the proposed classifier provides detailed information about the effect of different inputs on its final decisions. This makes our model intrinsically interpretable, which means that, 
unlike the model-agnostic explainability techniques it provides direct access to its internal decision-making mechanism. We applied the new approach to nine benchmark data sets, including both images and texts. 

For textual data, we demonstrated that the performance of AChorDS-LVQ in document classification task (with documents of varying lengths), is comparable (or better) than that of the transformer-based models, i.e. BERT, RoBERTa and Longformer. Additionally, we show that this new approach is more compute and resource-efficient since it follows the NPC strategy. The limited computing resource requirement of our proposed methodology also makes it an energy-efficient, environment-friendly, and affordable choice.

For image classification, we compared its performance with other prototype-based approaches. The experiments demonstrate the effectiveness of AChorDS-LVQ in image classification, achieving strong performance across multiple datasets.
Additionally, the model’s explainability is showcased through heatmap visualizations, which highlight the most important regions of an image that contribute to the classification decision. This provides valuable insights into how the model makes decisions and emphasizes its interpretability, making it easier to investigate and trust the model’s behavior. These findings underscore the potential of AChorDS-LVQ as a powerful and interpretable tool for image classification tasks.

Finally, for image-set classification tasks, even though the AChorDS-LVQ achieves similar performance to GRLGQ, it offers a significant advantage by providing insights into its decision-making. It highlights the images and specific pixels that influenced its classification decisions most.

In both image(-set) and document classification tasks, the newly proposed model enables detection of sources of uncertainties and bias
affecting its decision-making, and estimation of the strength of this uncertainty, which makes it a robust, and explainable model. These characteristics allow the end-users of this model to verify if its decision-making mechanism is sensible and fair, thereby making it reliable
for anthropocentric applications.

\section*{Acknowledgment}
This research is part of the research project \emph{EVICT} (\url{www.eviction.eu}), funded the the European Union’s ERC Research Grant under grant agreement No 949316, which supported M.M. S.G. has been supported by ITEA4 DAIsy 21016 (\url{https://itea4.org/project/daisy.html}). 

\bibliographystyle{unsrt}  
\bibliography{references}

\appendix
\clearpage
\renewcommand\thetable{\thesection.\arabic{table}}   
\setcounter{table}{1} 
\setcounter{page}{1}
\setcounter{section}{0}
\renewcommand{\thepage}{A\arabic{page}}
\renewcommand{\thesection}{A.\ \arabic{section}}  
\renewcommand{\thetable}{A.\ \arabic{table}}  
\renewcommand{\thefigure}{A.\ \arabic{figure}} 
\renewcommand{\theequation}{A.\ \arabic{equation}} 

\section{Appendix: Optimization} \label{subsec:gradient-computation}
To optimize the cost-function, we use SGD, requiring the computation of the first-order derivatives. Thus, in the following, we compute the (Euclidean) gradient of the cost-function Eq.\ \eqref{eq:adaptive_chordal}.  
Thereafter, one can use this gradient to estimate the Riemannian gradient of the cost-function (
details in \cite{edelman1998geometry}). 
Let $P$ be a randomly selected training example. Together, Eq.\ \eqref{eq:cost_function} and \eqref{eq:adaptive_chordal} show that the explained cost value only depends on the winner prototypes $W^{\pm}$ and the relevance factors $\lambda$, i.e., only these terms need to be updated. Thus, we calculate the derivative of $E_s$ with respect to these parameters. 
From the computation of $d_c(P, W^{\pm})$, we obtain principal vectors of $P$ and $W$, denoted as $U$ and $V$, respectively (see Eqs.\ \eqref{eq:compute_cc} and \eqref{eq:PQ}). As $V$ is just a rotation of $W$ (generating the same subspace), $W$ and $V$ represent the same point on the Grassmann manifold. This means that we can interchangeably compute the derivative with respect to $W$ and $V$. Thus, instead of $W$, we compute the derivative of the cost-function with respect to the principal vectors as follows:
\begin{align}
    \label{eq:derE_W}
    \frac{\partial E_s}{\partial V^{\pm}} &= \frac{\partial E_s}{\partial \mu} \frac{\partial \mu}{\partial d_c(P, V^{\pm})} \frac{\partial d_c(P, V^{\pm})}{\partial V^{\pm}} \enspace, 
    \\
    \frac{\partial \mu}{\partial d_c(P, V^{\pm})} &= \pm \frac{2  d_c(P, V^{\mp})}{\left( d_c(P, V^+) + d_c (P, V^-) \right)^2 } \enspace,\\
    \frac{\partial d_c(P, V)}{\partial \vect{v}_k} &\overset{\text{Eq.} 
 (\ref{eq:adaptive_chordal})}{=} - \lambda_k \vect{u}_k \enspace.
\end{align}
where 
\[
    \frac{\partial E_s}{\partial \mu} =
    \begin{cases}
         \mu (1 - \mu) & \text{if }\phi \text{ is sigmoid} \\
         1 & \text{if }\phi \text{ is identity}
    \end{cases}    
    \enspace,
\]

This results in the following gradient:
\begin{equation}
    \frac{\partial E_s}{\partial V^{\pm}} = \pm \frac{2  d_c(P, W^{\mp})}{\left( d_c(P, W^+) + d_c (P, W^-) \right)^2 } (UG)^{\pm}\enspace,
    \label{eq:deri_proto}
\end{equation}
where $G$ is a $(\dimension \times \dimension$)-diagonal matrix containing relevance factors.
Similarly, we can compute the derivative of the cost-function with respect to the relevance factors:
\begin{align}
    \label{eq:grad_lamda}
    \frac{\partial E_s}{\partial \lambda_k} = \frac{\partial E_s}{\partial \mu_i} &\left[ \frac{\partial \mu_i}{\partial d_c(P, V^{+})} \frac{\partial d_c(P, V^{+})}{\partial \lambda_k} + \frac{\partial \mu_i}{\partial d_c(P, V^{-})} \frac{\partial d_c(P, V^{-})}{\partial \lambda_k} \right], \enspace 
\end{align}
where 
\[
    \frac{\partial d_c^{\lambda}(P, V)}{\partial \lambda_k} = \vect{u}_k^T \vect{v}_k 
\]
Using the computed derivative in Eq.\ \ref{eq:deri_proto} and \ref{eq:grad_lamda}, we define the following updating rules:
\begin{align}
    \label{eq:update_W}
    W^{\pm} &\leftarrow V^{\pm} - \eta_{W^\pm} \frac{\partial E_s}{\partial V^{\pm}}\\
    \lambda &\leftarrow \lambda - \eta_{\lambda} \frac{\partial E_s}{\partial \lambda}
\end{align}

where $\eta_{W^\pm}$ and $\eta_{\lambda}$ are their respective learning rates. 
We set $\eta_{\lambda} \ll \eta_{W^\pm}$ as relevance factors will be updated (much) more often than prototypes. 
Note that after any update, we orthonormalize $W^{\pm}$\footnote{As we need orthonormalized matrices for computing canonical correlations.} and normalize $\{\lambda_i\}_{i=1}^\dimension$.

\section{Appendix: Experiments} \label{A-sec:dataset}

\subsection{Text Classification}
In this experiment, we fix learning rates as $\eta_W = 0.1, \eta_{\lambda} = 10^{-5}$, and use the sigmoid function for $\phi$ (in Eq.\ \eqref{eq:cost_function}).
For the first two datasets, consisting of shorter text, we set $\dimension = 20$, and for the last dataset (i.e. Arxiv-4), containing long texts, we fix $\dimension = 30$.

For our proposed model and the one conceptually similar to it, we have also provided the standard deviations over three random initializations. Since running transformer-based models hav been too compute-resource intensive, we have reported their performance for only one run at this moment.

\paragraph{Reuters-8:} It is a preprocessed version of the Reuters-21578 and consists of short texts categorized into eight classes: acq, crude, earn, grain, interest, money-fx, ship, and trade. It contains 5,485 training and 2,189 test examples.

\paragraph{Hyperpartisan:} This dataset \cite{kiesel-etal-2019-semeval} is labelled based on whether an article is hyperpartisan or not. 
It is quite different from other datasets in that it is a very small dataset, consisting of 516 documents for training and 64 and 65 documents as validation and test sets, respectively. 

\paragraph{Arxiv-4:} The third dataset is the Arxiv-4 dataset \cite{liu2018long}, comprising 12,195 research papers from the arXiv repository, classified into four distinct categories: cs.IT, cs.Ne, math.AC, and math.GR. In contrast to the Reuters-8 and the Hyperpartisan datasets, the Arxiv-4 dataset contains significantly longer documents, with an average length exceeding 6,000 words. Following \cite{liu2018long,he2019long}, we use 80\% of documents for training and the remaining for testing.

\subsection{Image classification}

\paragraph{Selected hyperparameters}
For the experiments within image classification, we set hyperparameters as stated in the table \ref{tab:hyperparameters_image_classification}. 
%
\begin{table}[]
    \centering
    \begin{tabular}{cccc}
        \hline
        hyperpara.& ResNet50 & ConvNeXt-Tiny & Definition\\
        \hline
        $\dimension$ & 10 & 10 & dim. of subspaces \\
        $\phi$  & Identity & Sigmoid & (see Eq. \eqref{eq:cost_function})\\
        $\eta_W$& $0.01$ & $0.05$ & learn. rate of prototypes  \\
        $\eta_{\lambda} $ & $10^{-5}$ & $10^{-6}$ &learn. rate of relevance \\
        $\eta_{\text{block}}$ & $10^{-4}$ & $10^{-4}$& learn. rate of the block\\
        $\eta_{\text{net}}$ & $10^{-6}$& $10^{-5}$ & learn. rate of the net\\
        $W\times H\times D$ & $7 \times 7 \times 512$& $13 \times 13 \times 512$& shape of the block\\
        \hline
    \end{tabular}
    \caption{The list of hyperparameters and their corresponding values (for image classification).}
    \label{tab:hyperparameters_image_classification}
\end{table}

\paragraph{CUB-200-2011:} The CUB-200-2011 dataset is a widely used benchmark for fine-grained image classification, containing 11,788 images of 200 bird species. Each class corresponds to a different bird species. The dataset is split into 5,994 training images and 5,794 test images, making it well-suited for evaluating classification models on fine-grained visual distinctions. 

\paragraph{Stanford Cars:} The Stanford Cars dataset is a collection of 16,185 images categorized into 196 distinct car classes, typically defined at the level of make, model, and year (e.g., Aston Martin Virage Convertible 2012). The dataset is evenly divided into 8,144 training images and 8,041 testing images, ensuring a balanced distribution across all classes. This dataset is widely utilized for fine-grained vehicle classification and recognition tasks.

\paragraph{Oxford-IIIT Pet:} The Oxford-IIIT Pet Dataset is a collection of 7,349 images featuring 37 distinct breeds of cats and dogs, with about 200 images per breed. The dataset is divided into training and testing subsets, with 3,680 images for training and 3,669 for testing. This dataset is widely utilized for tasks such as image classification, object detection, and semantic segmentation, serving as a benchmark for evaluating and comparing machine learning models in the field of computer vision.

\subsection{Image-set classification}

For this experiment, we set $\eta_W=0.1, \eta_{\lambda}=10^{-5}$, and we use the identity function as $\phi$ in Eq.\ \eqref{eq:cost_function}.
Furthermore, to maintain a fair comparison with GRLVQ, we set the dimensionality $\dimension$ to the values specified in \cite{mohammadi2024generalized}, which are 5, 25, and 22 for the ETH-80, YaleB, and UCF datasets, respectively.

\paragraph{Object Recognition:} The ETH-80 dataset, comprising 8 object categories, is widely utilized for object classification \cite{leibe2003analyzing}. Each category contains 10 image sets, with 41 images per set captured from various viewpoints. Following \cite{wei2020prototype}, images were converted to grayscale and resized to $20 \times 20$. For evaluation, data was split into training and testing sets, with five randomly chosen image sets per object for training. This process was repeated 10 times and then the mean and the standard deviations of accuracies. 

\paragraph{Face Recognition:} The Extended Yale Face Database B (YaleB) encompasses 16,128 images featuring 28 individuals captured across varied poses and illumination conditions (with 9 poses and 64 illumination conditions per person). Following the preprocessing steps outlined in \cite{wei2020prototype}, we organized the images (of size $20 \times 20$) into 9 sets per subject based on their poses. To create training and testing sets, three image sets were randomly chosen for training (per individual), with the remainder reserved for testing. This process was iterated ten times, after which the mean accuracy and standard deviation were computed.

\paragraph{Activity recognition:} The UCF Sports dataset consists of 150 videos of ten actions. They have been captured in different scenes and from different viewpoints \cite{rodriguez2008action}. To prepare videos for training, we followed the same preprocessing steps as \cite{sogi2020metric,mohammadi2024generalized}. 
Similar to \cite{sogi2020metric,mohammadi2024generalized}, we used a leave-one-out cross-validation scheme (LOOCV) to report the performance for the UCF data set.

\section{Appendix: Explainability}\label{A-sec:Explain}
As it has been shown in section \ref{subsec:explainability}, we can investigate the role of any input vector on the model's final prediction. Following Eq.\ \eqref{eq:explainability_diff}, we investigate the role of inputs on models' predictions for several examples from different classification tasks.

\subsection{Document classification}\label{A-subs:reuters}
\begin{figure}[h!]
    \begin{quotation}
    \scriptsize{
    \begin{mdframed}
    US oil dependency seen rising to record level the united states dependency on foreign oil sources may reach record levels by the mid s according to john h lichtblau president of petroleum industry research associates lichtblau speaking at an alternative energy conference here said the US may depend on foreign suppliers for as much as pct of its oil by surpasssing the previous high level of pct in the long term growth in dependency on foreign oil is inevitable lichtblau said as much as pct of US oil imports in could come from opec nations he said lichtblau said the US depended on foreign suppliers for pct of its oil in and predicted that would increase to pct by however the rate of this growth can be affected positively or negatively through government action or inaction lichtblau said he said that one of the government s negative actions is the maintenance of the windfall profits tax which acts as a disincentive to developing existing fields and reduces cash flow for oil exploration lichtblau called for the adoption of an international floor price for crude oil to help stabilize world oil prices an international floor price adopted by all or most industrial countries would clearly be a much more effective measure and would be much less distortive for the US than if we imposed it alone lichtblau said development of alternate energy sources such as synthetic fuels as well as increased development in alaska could lessen US dependency on foreign oil lichtblau said a potential for alternative supplies could limit the willingness of opec nations to raise oil prices he said lichtblau also called for the federal government to offer tax abatements for oil drilling to fill the strategic petroleum reserve at a faster rate and to develop pilot plans for alternative energy reuter.
    \end{mdframed}
    }
    \end{quotation}
    \caption{Example text from the `Crude' class in the Reuters-8 dataset, based on which \autoref{fig:topwords_retuers-8} is generated.}
    \label{fig:reuters_example}
\end{figure}

\begin{minipage}{\textwidth}
  \begin{minipage}[h!]{0.43\textwidth}
    \includegraphics[width=\textwidth]{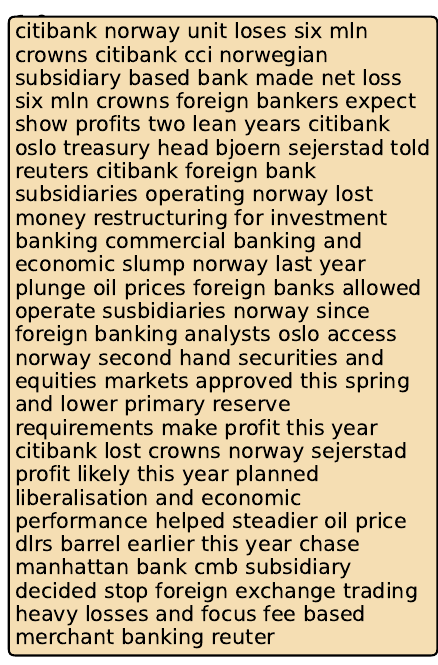}
    \captionof{figure}{A sample text from the `Earn' class in the Reuters-8 dataset.}
    \label{fig:reuters_example2}
  \end{minipage}
  \hfill
  \begin{minipage}[h]{0.53\textwidth}
    \centering
    \includegraphics[width=\textwidth]{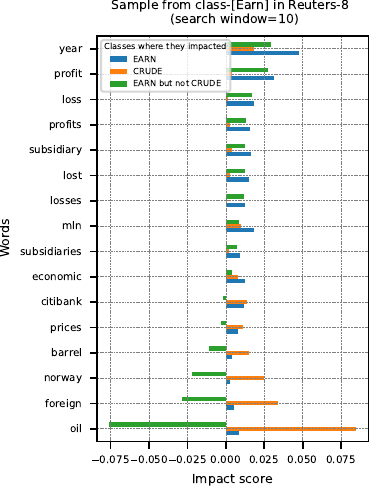}
    \captionof{figure}{Words with the highest impact on the identification of sample from \autoref{fig:reuters_example2} as \emph{Earn}.}
    \label{fig:reuter-explain2}
  \end{minipage}
\end{minipage}

\begin{minipage}{\textwidth}
  \begin{minipage}[h!]{0.5\textwidth}
    \includegraphics[width=\textwidth]{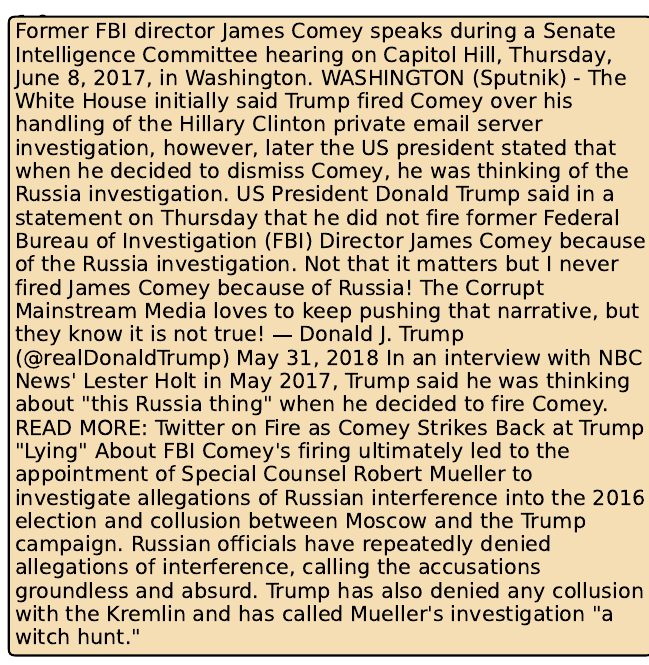}
    \captionof{figure}{An example text from the 'non-Hyperpartisan' class in the Hyperpartisan dataset.}
    \label{fig:hyperpartisan_example}
\end{minipage}
\hfill
 \begin{minipage}[h!]{0.49\textwidth}
    \centering
    \includegraphics[width=1.\textwidth]{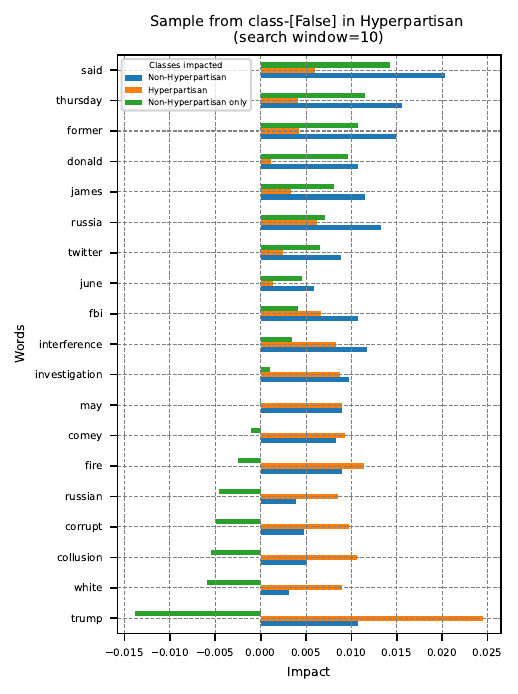}
    \captionof{figure}{Words with the highest impact on the identification of sample \autoref{fig:hyperpartisan_example} as \emph{non Hyperpartisan}.}
    \label{fig:hyperpartisan-explain}
\end{minipage}
\end{minipage}
\smallskip

\begin{minipage}{\textwidth}
  \begin{minipage}[h!]{0.5\textwidth}
    \hspace{-5pt}
    \includegraphics[width=1.\textwidth]{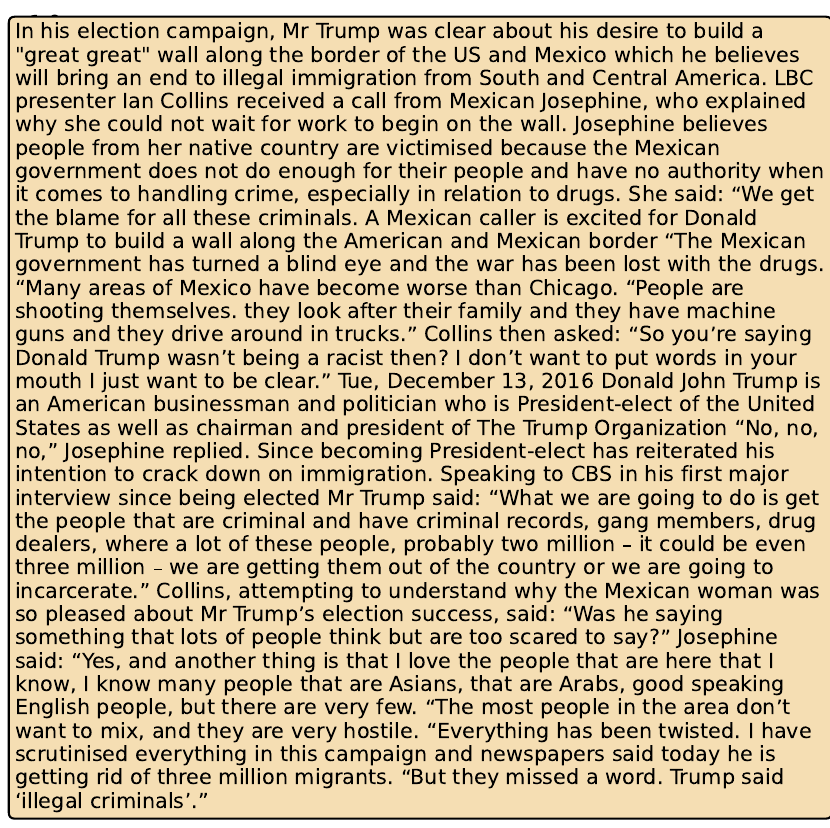}
    \captionof{figure}{An example text from the `Hyperpartisan' class in the Hyperpartisan dataset.}
    \label{fig:hyperpartisan_example2}
\end{minipage}
\hspace{5pt}
 \begin{minipage}[h]{0.45\textwidth}
   \hspace{-3pt} 
    \centering
    \includegraphics[width=1.15\textwidth]{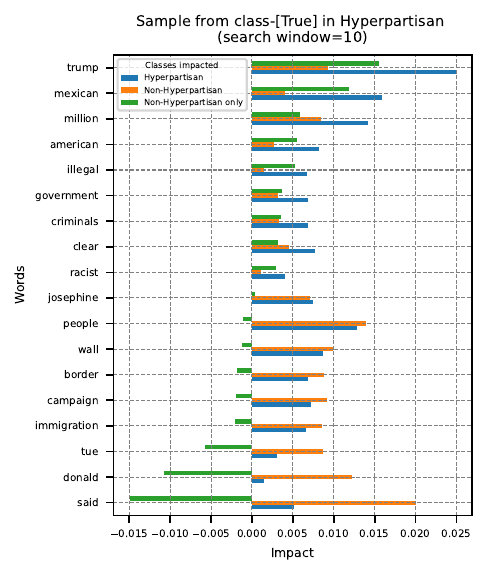}
    \captionof{figure}{Words with the highest impact on the identification of sample from \autoref{fig:hyperpartisan_example2} as \emph{Hyperpartisan}.}
    \label{fig:hyperpartisan-explain2}
\end{minipage}
\end{minipage}

\begin{minipage}{\textwidth}
  \begin{minipage}[h!]{0.47\textwidth}
    \centering
    \hspace{-15pt}
    \includegraphics[width=1.\textwidth]{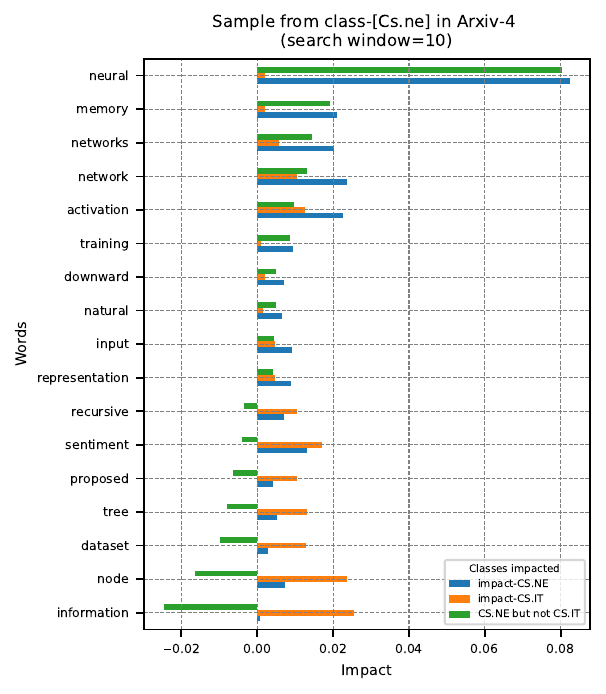}
    \captionof{figure}{Words with the highest impact on the identification of the paper \emph{\href{https://arxiv.org/abs/1701.01811}{Structural Attention Neural Networks for improved sentiment analysis}} as a sample from class \emph{CS.NE}, of the Arxiv-4 dataset.}
    \label{fig:arxiv-4-explain1}
\end{minipage}
 \begin{minipage}[h]{0.47\textwidth}
    \centering
    \hspace{10pt}
    \includegraphics[width=1.\textwidth]{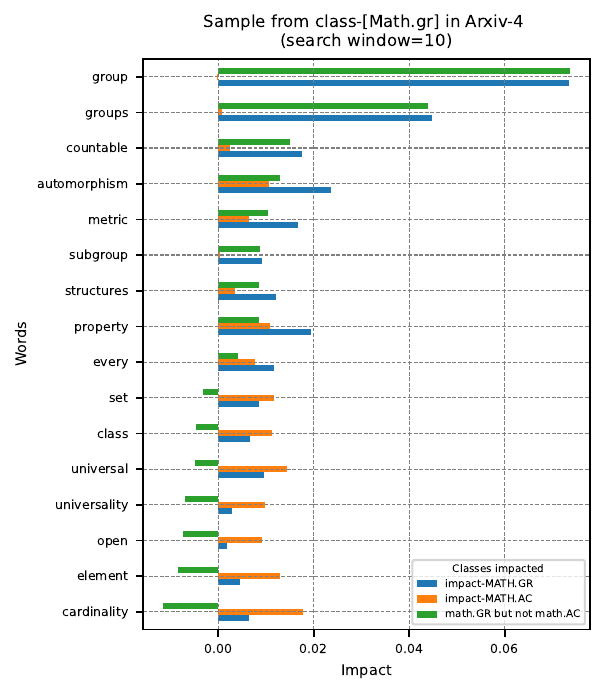}
    \captionof{figure}{Words with the highest impact on the identification the paper \emph{\href{https://arxiv.org/abs/1407.4727}{Non-universality of automorphism groups of uncountable ultrahomogeneous structures}} as a sample from class \emph{math.GR} of the Arxiv-4 dataset.}
    \label{fig:arxiv-4-explain2}
\end{minipage}
\end{minipage}

\subsection{Image classification}

\begin{figure}
    \centering
    \includegraphics[width=0.3\linewidth]{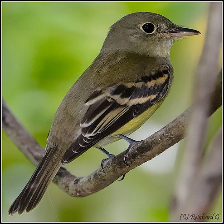}
    \includegraphics[width=0.3\linewidth]{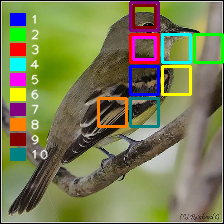}
    \includegraphics[width=0.3\linewidth]{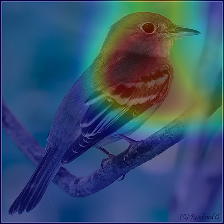}
    \caption{Left) An example image of an Acadian Flycatcher from the CUB-200-2011 dataset. 
    Middle: The highlighted regions per principal direction. 
    Right) The combined images comprised both the image and the heatmap generated by our model. 
    The backbone network used is ResNet50.}
    \label{fig:indigo_bunting}
\end{figure}

\begin{figure}
    \centering
    \includegraphics[width=0.3\linewidth]{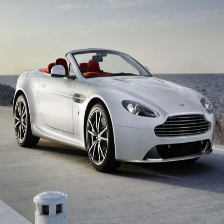}
    \includegraphics[width=0.3\linewidth]{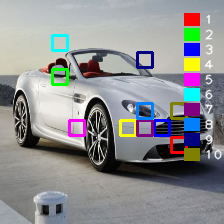}
    \includegraphics[width=0.3\linewidth]{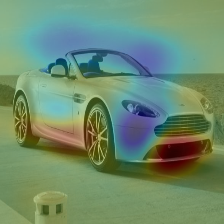}
    \caption{Left) An example image of an Aston Martin V8 Vantage Convertible 2012 from the Cars dataset. 
    Middle: The highlighted regions per principal direction. 
    Right) The combined images comprised both the image and the heatmap generated by our model.
    The backbone network used is ConvNeXt-Tiny.}
    \label{fig:cars}
\end{figure}

\begin{figure}
    \centering
    \includegraphics[width=0.3\linewidth]{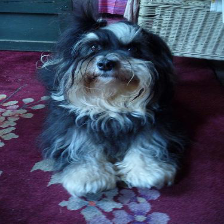}
    \includegraphics[width=0.3\linewidth]{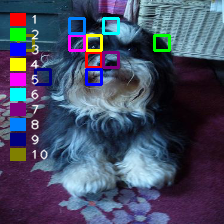}
    \includegraphics[width=0.3\linewidth]{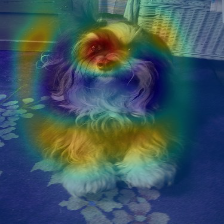}
    \caption{Left) An example image of an Havanese from the Pets dataset. 
    Middle: The highlighted regions per principal direction. 
    Right) The combined images comprised both the image and the heatmap generated by our model.
    The backbone network used is ConvNeXt-Tiny.}
    \label{fig:havanese}
\end{figure}

\subsection{Image-set classification}\label{A-subs:ETH80}
\begin{figure}[h!]
    \centering
    \textbf{(a)}
    \includegraphics[scale=0.88]{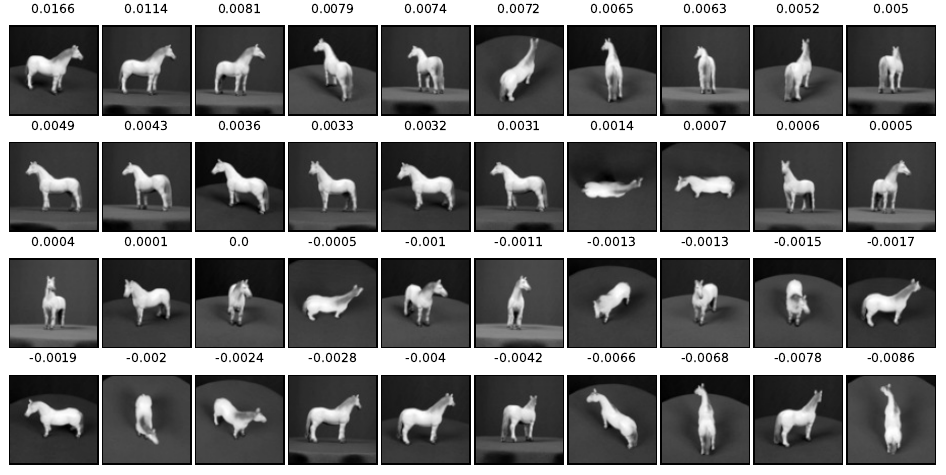}
    \textbf{(b)}
    \includegraphics[scale=0.89]{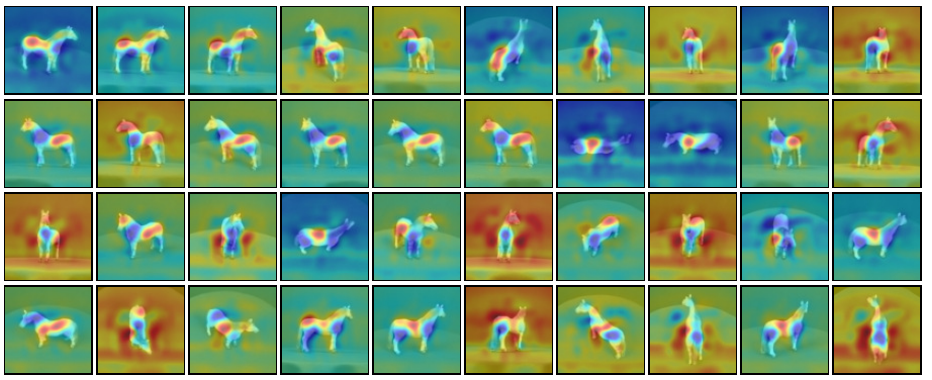}
    \caption{a) Sorted images based on their impact on the model decision. b) Highlighted pixels for sorted images.}
    \label{fig:example-eth80}
\end{figure}

Given an image set of a horse (from the ETH-80 dataset), the model could correctly classify it. \autoref{fig:example-eth80}a shows the sorted images based on their impact where their impact is above figures. \autoref{fig:example-eth80}b displays heatmaps capturing the influence of pixels within each image.

\section{Appendix: Model and Compute resource specifications}\label{A-sec:Compute resource}
\def\myColW{0.2\textwidth}
 \begin{table}[h!]
    \setlength\tabcolsep{1pt}
     \caption{The performance of different classifiers on document classification task.}\label{tab:doc_classification_perf}
     \begin{tabularx}{\linewidth}{@{\extracolsep{\fill}} 
            @{}>{\raggedright\footnotesize}p{0.16\textwidth}@{}*{2}{@{}>{\raggedright\footnotesize}p{\myColW}@{}}
            >{\footnotesize\arraybackslash}l@{} 
            }
\toprule
        \textbf{Model} & \textbf{Version and documentation link} &\textbf{Computing resource used} & \makecell{\textbf{$\approx$ Compute time}}\\
\toprule
BERT&  \href{https://huggingface.co/google-bert/bert-base-uncased}{google-bert/bert-base-uncased }& T4GPU (via Google Collaboratory) & 8.41min (hyperpartisan), 171min(arxiv-4)\\
RoBERTa & \href{https://huggingface.co/FacebookAI/roberta-base}{FacebookAI/roberta-base} & T4GPU (via Google Collaboratory) & 9.20min (hyperpartisan), 179min(arxiv-4)\\
Longformer& \href{ttps://huggingface.co/allenai/longformer-base-4096}{allenai/longformer-base-4096}& A100GPU (via Google Collaboratory) & 301min(arxiv-4)\\
word2vec& \href{https://radimrehurek.com/gensim/models/word2vec.html}{word2vec-google-news-30} & & \\
GloVe & \href{https://nlp.stanford.edu/projects/glove/}{glove.42B.300d} & & \\
SVM & \href{https://scikit-learn.org/stable/modules/generated/sklearn.svm.SVC.html}{Sklearn SVM}&  6 cores of CPU & \\
GRLGQ & \href{https://github.com/mohammadimathstar/GRLGQ:}{Code repository of GRLGQ in Github}& 6 cores of CPU & 55 second(hyperpartisan), 81min(arxiv-4)\\
AChorDS-LVQ & Proposed method in this contribution. Code repository will be made public soon after acceptance of this work. & 6 core of CPU & 50 second (hyperpartisan) , 76min (arxiv-4)\\
\bottomrule
\addlinespace[5pt]
  \end{tabularx}
\end{table}

\medskip

\end{document}